\documentclass[12pt,authoryear,preprint]{elsarticle}

\usepackage{url}

\usepackage[utf8]{inputenc}
\usepackage[english]{babel} 
\usepackage{amsmath,amsthm, amssymb} 
\usepackage[mathscr]{euscript}
\usepackage{soul}
\usepackage{algorithmicx}
\usepackage[ruled]{algorithm}
\usepackage{algpseudocode}
\usepackage{graphicx}
\usepackage{xcolor}
\usepackage{tabularx}

\makeatletter
\def\ps@pprintTitle{%
 \let\@oddhead\@empty
 \let\@evenhead\@empty
 \def\@oddfoot{\centerline{\thepage}}%
 \let\@evenfoot\@oddfoot}
\makeatother
\newtheorem{theorem}{Theorem}

\begin{document}
\graphicspath{{./figs/}}

\begin{frontmatter}

\title{Deep Deterministic Policy Gradient\\ for Urban Traffic Light Control}

\author{Noe Casas}
\ead{contact@noecasas.com}

\journal{Engineering Applications of Artificial Intelligence}

\begin{abstract}

Traffic light timing optimization is still an active
line of research despite the wealth of scientific
literature on the topic, and the problem remains
unsolved for any non-toy scenario. One of the key
issues with traffic light optimization is the large
scale of the input information that is available for the
controlling agent, namely all the traffic
data that is continually sampled by the traffic detectors
that cover the urban network. This issue has in the past forced
researchers to focus on agents that work on localized
parts of the traffic network, typically on
individual intersections, and to coordinate
every individual agent in a multi-agent setup.
In order to overcome the large scale of the available
state information,
we propose to rely on the ability of deep Learning
approaches to handle large input spaces,
in the form of Deep Deterministic Policy Gradient (DDPG)
algorithm. We performed several experiments with a range of models,
from the very simple one (one intersection) to the more
complex one (a big city section).

\end{abstract}

\begin{keyword}
deep learning \sep reinforcement learning \sep traffic light control
\sep policy gradient
\end{keyword}

\end{frontmatter}

\section{Introduction}

Cities are characterized by
the evolution of their transit dynamics. Originally
meant solely for pedestrians, urban streets soon shared usage with
carriages and then with cars. Traffic organization became soon
an issue that led to the introduction of signaling, traffic lights
and transit planning.\\

Nowadays, traffic lights either have fixed programs or are actuated.
Fixed programs (also referred to as \textit{pretimed control}) are
those where the timings of the traffic lights
are fixed, that is, the sequences of red, yellow and green phases
have fixed duration. Actuated traffic lights change their
phase to green or red depending on traffic detectors that are
located near the intersection; this way, actuated
traffic light are dynamic and adapt to the traffic conditions to
some degree; however, they only take into account the conditions
local to the intersection. This also leads to dis-coordination with
the traffic light cycles of other nearby intersections and hence
are not used in dense urban areas.
Neither pretimed or actuated traffic lights take into account the current
traffic flow conditions at the city level. Nevertheless, cities have
large vehicle detector 
infrastructures that feed traffic volume forecasting tools used 
to predict congestion situations. Such information is normally only used
to apply classic traffic management actions like sending police officers
to divert part of the traffic.\\

This way, traffic light timings could be improved by means of machine learning
algorithms that take advantage of the knowledge about traffic conditions
by optimizing the flow of vehicles.
This has been the subject of several lines of research in the past.
For instance, Wiering proposed different variants
of reinforcement learning to be applied  to traffic light control
\citep{wiering2004simulation},
and created the \textit{Green Light District} (GLD) simulator
to demonstrate them,
which was further used in other works like \citep{prashanth2011reinforcement}.
Several authors explored the feasibility of applying fuzzy logic,
like \citep{favilla1993fuzzy} and \citep{chiu1993adaptive}. Multi-agent
systems where also applied to this problem, like \citep{cai2007study}
and \citep{shen2011agent}.\\
 
Most of the aforementioned approaches simplify the scenario to a
single intersection or a reduced
group of them. Other authors propose multi-agent systems where each agent
controls a single intersection and where
agents may 
communicate with each other to share information to improve coordination
 (e.g. in a \textit{connected vehicle} setup \citep{feng2015real}) or
may receive a piece of shared information to be aware of the
crossed effects on other agents' performance
(\citep{el2013multiagent}). 
However, none of the aforementioned approaches fully profited from
the availability of \textit{all} the vehicle flow information, that is,
the decisions taken by those agents were in all cases partially informed.
The main justification for the lack of \textit{holistic} traffic light
control algorithms is the poor scalability of most algorithms. In
a big city there can be thousands of vehicle detectors and tenths of
hundreds of traffic lights. Those numbers amount for huge space
and action spaces, which are difficult to handle by
classical approaches.\\

This way, the problem addressed in this works is the devisal of an agent
that receives traffic data and, based on these, controls the traffic
lights in order to improve the flow of traffic, doing it at a large
scale.

\section{Traffic Simulation} \label{sec:trafficconcepts}

In order to evaluate the performance of our work, we make use
of a traffic simulation.
The base of a traffic simulation is the \textbf{network}, that is,
the representation of roads and intersections where the vehicles
are to move. Connected to some roads, there are \textbf{centroids},
that act as sources/sinks of vehicles. The amount of vehicles
generated/absorbed by centroids is expressed in a
\textbf{traffic demand matrix}, or origin-destination (OD) matrix,
which contains one cell per each
pair of origin and destination centroids. During a simulation,
different OD matrices can be applied to different periods of time
in order to mimic the dynamics of the real traffic through time.
In the roads of the network, there can be \textbf{traffic detectors},
that mimic induction loops beneath the ground that are able to
measure traffic data as vehicles go pass through them. Typical
measurements that can be taken with traffic detectors include
vehicle counts, average speed and percentage of occupancy.
There can also be traffic lights. In many cases they are used to
regulate the traffic at intersections. In those cases, all the
traffic lights in an intersection are coordinated so that
when one is red, another one is green, and vice versa
(this way, the use of the intersection is regulated so that
vehicles don't block the intersection due to their intention
to reach an exit of the intersection that is currently in use)
. All the traffic lights in the intersection
change their state at the same time. This intersection-level
configuration of the traffic lights is called a \textbf{phase}, and
it is completely defined by the states of each traffic light in the
intersection plus its duration. The different phases in an
intersection form its \textit{control plan}. The phases in the
control plan are applied cyclically, so the phases
are repeated after the \textit{cycle duration} elapses. Normally,
control plans of adjacent intersections are synchronized to
maximize the flow of traffic avoiding unnecessary stops.\\

Urban traffic simulation software can keep models at different
levels of abstraction. \textbf{Microscopic simulators} simulate
vehicles individually computing their
positions at every few milliseconds and the
the dynamics of the vehicles are governed by a
simplified model that drives the
behaviour of the driver under different conditions, while
\textbf{macroscopic simulators}
work in an aggregated way, managing traffic
like in a flow network in fluid dynamics. There are different
variations between microscopic and macroscopic models, broadly referred
to as \textbf{mesoscopic simulators}.
To our interests, the proper simulation level would be microscopic,
because we need information of individual vehicles and their
responses to changes in the traffic lights, mimicking closely
real workd dynamics in terms of congestion. As third party
simulator we chose \textbf{Aimsun} \citep{casas2010traffic, aimsun2012dynamic},
a commercial microscopic, mesoscopic and macroscopic simulator
widely used, both in the private consulting
sector and in traffic organization institutions.

\section{Preliminary Analysis}

The main factor that has prevented further advance in the
traffic light timing control problem is the large scale of
any realistic experiment. On the other hand, there is a
family of machine learning algorithms whose very strenght
is their ability of handle large input spaces, namely
\textbf{deep learning}.
Recently, deep learning has been successfully applied to reinforcement
learning, gaining 
much attention due to the effectiveness of Deep Q-Networks (DQN) 
at playing Atari games using as input the raw pixels of the game
\citep{mnih2013playing, mnih2015human}. Subsequent successes
of a similar approach called Deep Deterministic Policy
Gradient (DDPG) were achieved in 
\citep{lillicrap2015continuous}, which will be used in our
work as reference articles, given the similarity of the
nature of the problems addressed there, namely large continuous
state and action spaces.
This way, the theme of this work is the \textbf{application of
Deep Reinforcement Learning
to the traffic light optimization problem} with an holistic
approach, by leveraging deep learning to cope with the large
state and action spaces. Specifically, the \textbf{hypothesis} that
drives this work is that \textit{Deep reinforcement learning
can be successfully applied to urban
traffic light control, having similar or better performance than other
approaches}.\\

This is hence the main contribution of the present work, along
with the different techniques applied to make this application possible
and effective.
Taking into account the nature of the problem and the abundant
literature on the subject, we know that 
some of the challenges of devising a traffic light timing control
algorithm that acts at a large scale are:

\begin{itemize}
\item Define a sensible \textbf{state space}. This includes
finding a suitable representation of the traffic information. Deep learning
is normally used with input signals over which convolution is easily computable,
like images (i.e. pixel matrices) or sounds (i.e. 1-D signals). Traffic information
may not be easily represented as a matrix, but as a labelled graph.
This is addressed in section \ref{sec:inputrepresentation}.

\item Define a proper \textbf{action space} that our agent is able
to perform. The naive approach would be to let the controller simply control
the traffic light timing directly (i.e. setting the color of each traffic light
individually at each simulation step). This, however, may lead to
breaking the normal routing
rules, as the traffic lights in an intersection have to be synchronized
so that the different intersection exit routes do not interfere with
each other. Therefore a careful definition of the agent's actions
is needed. This is addressed in section \ref{sec:actions}.

\item Study and ensure the \textbf{convergence} of the
approach: despite the successes
of Deep Q-Networks and DDPG, granted by their numerous contributions
to the stability of reinforcement learning with value function
approximation, convergence of such approaches is not guaranteed.
Stability of the training is studied and measures for palliating
divergence are put in place. This is addressed in section
\ref{sec:gammaschedule}.

\item Create a sensible \textbf{test bed}: a proper test bed should simulate
relatively realistically the traffic of a big city, including a realistic
design of the city itself. This is addressed in section
\ref{chap:experiments}.

\end{itemize}

\section{Related Work} \label{chap:related}

In this section we identify and explore other lines of research
that also try to solve the traffic light control problem.

\subsection{Offline Approaches}

The most simple traffic light control approaches are those
that define \textbf{fixed} timings for the different
traffic light phases. These timings are normally defined
offline (i.e. not in closed loop). Several different 
approaches have been proposed in the literature for
deriving the phase timings, which can be grouped into the
following categories \footnote{The categorization focuses
on both the adaptative nature (or lack thereof) of the approach
and the type of algorithms used and their similarity to the
approach proposed in this work.}:
\begin{itemize}
\item Model-based: a mathematical model of the target urban
area is prepared and then used to derive an optimal timing,
either via derivative calculus, numerical optimization, 
integer linear programming, or any other method. An example  of
this approach is MAXBAND \citep{little1966synchronization}, which
defines a model for arterials and optimizes it for
maximum bandwidth by means of linear programming. Another
example is the TRANSYT system \citep{robertson1969transyt},
which uses an iterative process to minimize the average journey
time in a network of intersections.

\item Simulation-based: this case is analogous to the model-based,
but the core of the validation of the timings is a traffic
simulation engine, which is connected to a black box
optimization computation that iteratively searches the
traffic light timing space to find an optimal control
plan. Some examples of this approach are
\citep{rouphail2000direct}, which make use of genetic algorithms together
with the CORSIM simulator \citep{holm2007traffic}, or
\citep{garcia2013optimal}, which uses particle swarm optimization
with the SUMO simulator \citep{SUMO2012}.

\end{itemize}

The usual way of
maximizing the success of this kind of methods is 
to analyze historical traffic data and identify time
slots with different traffic volume characteristics;
once defined, a different timing strategy is derived
for each these \textit{time bands}. However, not even
this partitioning scheme adapts to the dynamism of
traffic demand or progressive changes in drivers'
behaviour.

\subsection{Model-based Adaptive Approaches}

The simplest of these approaches only one intersection
into consideration. They define a model (e.g. based
on queue theory) that is fed with real detector
data (normally from the closest detectors to the
intersection). Then, by using algorithmic logic
based on thresholds and rules, like \citep{lin89binary},
or optimization
techniques like \citep{shao2009adaptive}, they try to minimize
waiting times.\\

More complex approaches are based on traffic network models
of several intersections 
that are fed with the real time data from multiple traffic
detectors.  
Some of these approaches are heuristically defined algorithms
that \textit{tune} their parameters by performing
tests with variations on the aforementioned models.
For example, the SCOOT system \citep{hunt1982scoot}
performs small reconfigurations (e.g. individual intersection
cycle offsets or cycle splits) on a traffic network
model. More recent approaches like
\citep{tubaishat2007adaptive} make use of the information
collected by Wireless Sensor Networks (WSN) to pursue
the same goal.  
There are also approaches where more formal
optimization methods are employed on the traffic network
models fed with read time data, like the case of 
\citep{gartner1983opac}, \citep{henry1983prodyn},
\citep{boillot1992optimal} or \citep{sen1997controlled},
which compute in real time the switch times of
the traffic lights within the next following minutes
by solving dynamic optimization problems on
realistic models fed with data from real traffic
detectors.

\subsection{Classic Reinforcement Learning}

Reinforcement Learning has been applied in the past to urban traffic
light control. Most of the instances from the literature consist of a 
classical algorithm like Q-Learning, SARSA or TD($\lambda$) to control
the timing of a single intersection. Rewards are typically
based on the reduction of the travel time of the vehicles or the queue
lengths at the traffic lights. \citep{el2014design} offers a thorough review of
the different approaches followed by a subset of articles from the literature
that apply reinforcement learning to traffic light timing control. As
shown there, many studies use as state space information such as
the length of the queues and the travel time delay; such type of
measures are rarely available in a real-world setup and can therefore
only be obtained in a simulated environment. Most of the approaches
use discrete actions (or alternatively, discretize the continuous
actions by means of tile coding,
and use either $\varepsilon$-greedy selection (choose the action with
highest Q value with $1 - \varepsilon$ probability, or random action
otherwise) or softmax selection (turn Q values into probabilities
by means of the softmax function and then choose stochastically accordingly).
In most of the applications of reinforcement learning to traffic control,
the validation scenario consists of a single intersection, like in
\citep{thorpe1997vehicle}. This is due to
the scalability problems of classical RL tabular approaches: as the number
of controlled intersections increases, so grows the
state space, making the learning unfeasible due to the impossibility
for the agent to apply every action under every possible state.
This led some researchers to study multi-agent approaches, with varying
degrees of complexity: some approaches like that from
\citep{arel2010reinforcement} train each agent separately,
without notion that more agents even exist, despite the
coordination problems that this approach poses. Others like \citep{wiering2000multi}
train each agent separately, but only the intersection with maximum
reward executes the action.
More elaborated approaches, like in \citep{camponogara2003distributed},
train several agents together
modeling their interaction as a competitive stochastic game.
Alternatively, some lines of research like \citep{kuyer2008multiagent}
and \citep{bakker2010traffic}
study cooperative interaction of agents by means of coordination mechanisms,
like coordination graphs (\citep{guestrin2002coordinated}).\\

As described throughout this section, there are several examples in the
literature of the application of classical reinforcement learning to traffic
light control. Many of them focus on a single intersection. Others
apply multi-agent reinforcement learning techniques to address the
problems derived from the high dimensionality of state and action
spaces. Two characteristics of most of the explored approaches
are that the information used to elaborate the state space is hardly
available in a real-world environment and that there are no realistic
testing environments used.

\subsection{Deep Reinforcement Learning}

There are some recent works that, like ours, study the applicability of 
\textbf{deep} reinforcement learning to traffic light control:\\

Li et al. studied in \citep{li2016traffic} the application of deep
learning to traffic light timing in a single intersection.
Their testing setup consists of a single cross-shape intersection
with two lanes per direction, where no turns are allowed at all (i.e.
all traffic either flows North-South (and South-North) 
or East-West (and West-East), hence the traffic light set only
has two phases. This scenario is
therefore simpler than our simple network A presented in \ref{sec:simplea}.
For the traffic simulation, they use the proprietary software
PARAllel MICroscopic Simulation (Paramics) \citep{cameron1996paramics},
which implements the model by Fritzsche \citep{fritzsche1994model}.
Their approach consists of a Deep Q-Network
(\citep{mnih2013playing, mnih2015human}) comprised of a heap of
stacked auto-encoders \citep{bengio2007greedy, vincent2010stacked},
with sigmoid activation functions
where the input is the state of the network and the output is the
Q function value for each action.
The inputs to the deep Q network are the queue lengths of each lane at
time $t$ (measured in meters), totalling 8 inputs.
The actions generated by the network are 2: remain in the current
phase or switch to the other one. The reward is the absolute value
of the difference between the maximum North-Source flow and the maximum
East-West flow.
The stacked autoencoders are pre-trained (i.e. trained
using the state of the traffic as both input and output) layer-wise
so that an internal representation of the traffic state is learned, which
should improve the stability of the learning in further fine tuning
to obtain the Q function as output (\citep{erhan2010does}).
The authors use an experience-replay memory to improve learning convergence.
In order to balance exploration and exploitation, the authors use
an $\epsilon$-greedy policy, choosing a random action with a small
probability $p$.
For evaluating the performance of the algorithm, the authors
compare it with normal Q-learning (\citep{sutton1998reinforcement}).
For each algorithm, they show the queue lengths over time and perform
a linear regression plot on the queue lengths for each
direction (in order to check the \textit{balance} of their queue length).\\

Van der Pol explores in \citep{van2016deep}
the application of deep learning to traffic light coordination,
both in a single intersection and in a more complex configuration.
Their testing setup consists of a single cross-shaped intersection with
one lane per direction, where no turns are allowed.
For the simulation software, the author uses SUMO (Simulation of Urban MObility),
a popular open-source microscopic traffic simulator.
Given that SUMO \textit{teleports} vehicles that have been stuck for a long time
\footnote{See \url{http://sumo.dlr.de/wiki/Simulation/Why_Vehicles_are_teleporting}},
the author needs to take this into account in the reward function, in order
to penalize traffic light configurations that favour vehicle teleportation.
Their approach consists on a Deep Q-Network. The author experiments with two
two alternative architectures, taken verbatim respectively from
\citep{mnih2013playing} and \citep{mnih2015human}. Those convolutional networks were
meant to play Atari games and receive as input the pixel matrix with
bare preprocessing (downscaling and graying). In order to enable those
architectures to be fed with the traffic data as input, an image is created by plotting
a point on the location of each vehicle. The action space is comprised of
the different legal traffic light configurations (i.e. those that do not lead
to flow conflicts), among which the network chooses which to apply.
The reward is a weighted sum of several factors: vehicle delay (defined as
the road maximum speed minus the vehicle speed, divided by the road maximum
speed), vehicle waiting time, the number of times the vehicle stops, the
number of times the traffic light switches and the number of \textit{teleportations}.
In order to improve convergence of the algorithm, the authors apply
deep reinforcement learning techniques such as prioritized
experience replay and keeping a shadow target network, but also
experimented with double Q learning \citep{hasselt2010double, van2015deep}.
They as well tested different
optimization algorithms apart from the normal stochastic gradient optimization,
such as the ADAM optimizer \citep{kingma2014adam},
Adagrad \citep{duchi2011adaptive} or RMSProp \citep{tieleman2012lecture}.
The performance of the algorithm is evaluated visually by means
of plots of the reward and average travel time during the training phase.
The author also explores the behaviour of their algorithm in a scenario
with multiple intersections (up to four) by means of a multi-agent approach.
This is achieved by training two neighbouring intersections on their mutual
influence and then the learned joint Q function is \textit{transferred}
for higher number of intersections.\\

Genders et al. explore in \citep{genders2016using} the 
the application of deep convolutional learning to traffic light
timing. Their test setup consists of a single cross-shaped intersection
with four lanes in each direction, where the inner lane is meant
only for turning left and the outer lane is meant only for turning
right. As simulation software, the authors use SUMO, like the work
by Van der Pol \citep{van2016deep} (see previous bullet). However, Genders
et al do not address the teleportation problem and do not take into account
its effect on the results.
Their approach consists of a Deep Convolutional Q-Network. Like in
\citep{van2016deep}, Genders et al. transform the vehicle positions
into a matrix so that it becomes a suitable input for the convolutional
network. They, however, scale the value of the pixels with the local
density of vehicles. The authors refer to this representation as
discrete traffic state encoding (DTSE).
The actions generated by the Q-Network are the
different phase configurations of the traffic light set in the
intersection. The reward defined as the variation in cumulative vehicle
delay since the last action was applied.  
The network is fed using experience replay.

\section{Theoretical Background} \label{chap:background}

\textit{\textbf{Reinforcement Learning}} (RL) aims at
training an agent so that it applies actions optimally to an \textit{environment}
based on its state,
with the downside that it is not known which actions are good or bad, but
it is possible to evaluate the goodness of their effects after they are applied.
Using RL terminology, the goal of the algorithm is to learn an optimal policy
for the agent, based on the observable state of the environment and on a
\textit{reinforcement signal} that represents the reward (either positive
or negative) obtained when an action has been applied.
The underlying problem that reinforcement learning tries to solve is
that of the \textit{credit assignment}. For this, the algorithm normally
tries to estimate the expected cumulative future reward to be obtained
when applying certain action when in certain state of the environment.
RL algorithms act at discrete points in time. At each time step $t$, the agent tries to maximize
the expected total return $R_T$, that is, the accumulated rewards obtained after each
performed action: $R_t = r_{t+1} + r_{t+2} + \cdots + r_T$, where $T$ is the
number of time steps ahead until the problem finishes. However, as normally $T$ is
dynamic or even infinite (i.e. the problem has no end), instead of the
summation of the rewards,the discounted return is used:
\begin{equation} \label{eq:discounted}
R_t = r_{t+1} + \gamma r_{t+2} + \gamma^2 r_{t+3} + \cdots = \sum_{k=0}^{\infty} \gamma^k r_{t+k+1}
\end{equation}
The state of the environment is observable, either totally
or partially. The
definition of the state is specific to each problem. One example of state of the
environment is the position $x$ of a vehicle that moves in one dimension. Note
that the state can certainly contain information that condenses pasts states
of the environment. For instance, apart from the position $x$ from the previous
example, we could also include the speed $\dot{x}$ and acceleration $\ddot{x}$
in the state vector.
Reinforcement Learning problems that depend only on the current state
of the environment are said to comply with the \textit{Markov property} and
are referred to as \textit{Markov Decision Processes}. Their dynamics are
therefore defined by
the probability of reaching from a state $s$ to a state $s'$ by means of
action $a$:
\begin{equation}
p(s'|s,a) = P(S_{t+1}=s' | S_t=s, A_t=a)
\end{equation}
This way, we can define the reward obtained when transitioning from
state $s$ to $s'$ by means of action $a$:
\begin{equation}
r(s,a,s')= \mathbb{E} \left[ R_{t+1} | S_t = s, A_t = a, S_{t+1} = s' \right]
\end{equation}
\textbf{Deep Reinforcement Learning} refers to reinforcement learning algorithms
that use a deep neural network as value function approximator.
The first success of reinforcement learning with neural networks as
function approximation was TD-Gammon \citep{tesauro1995temporal}. Despite
the initial enthusiasm in the scientific community, the approach
did not succeed when applied to other problems, which led to its abandonment
(\citep{pollack1997did}). The main reason for its failure
was \textbf{lack of stability} derived from:
\begin{itemize}
\item The neural network was trained with the values that were
generated \textit{on the go}, therefore such values were sequential
in nature and thus were \textbf{auto-correlated}
(i.e. not independently and identically distributed).
\item \textbf{Oscillation of the policy} with small changes to Q-values
that change the data distribution.
\item Too large optimization steps upon \textbf{large rewards}.
\end{itemize}
Their recent rise in
popularity is due to the success of Deep Q-Networks (DQN) 
at playing Atari games using as input the raw pixels of the game
\citep{mnih2013playing, mnih2015human}.
\begin{equation} \label{eq:dqnloss}
\mathscr{L}(\theta) =
\mathbb{E} \left[ \left(y - Q(s,a ; \theta) \right)^2 \right]
\end{equation}
In DQNs, there is a neural network that receives the environment
state as input and generates as output the Q-values for each of
the possible actions, using the loss function \eqref{eq:dqnloss},
which implies following the direction of the gradient \eqref{eq:dqngradient}:
\begin{equation} \label{eq:dqngradient}
\nabla_\theta \mathscr{L}(\theta)=
\mathbb{E} \bigl[
\bigl( r + \gamma \max_{a'} Q(s',a' ; \theta)
- Q(s,a ; \theta) \bigr)
\nabla_\theta Q(s, a; \theta)
\bigr]
\end{equation}
In order to mitigate the stability problems inherent
to reinforcement learning with value function approximation, in
\citep{mnih2013playing,mnih2015human},
the authors applied the following measures:
\begin{itemize}
\item \textbf{Experience replay}: keep a memory of past action-rewards and train
the neural network with random samples from it instead of using the
real time data, therefore eliminating the temporal autocorrelation problem.
\item \textbf{Reward clipping:} scale and clip the values of the rewards to
the range $[-1, +1]$ so that the weights do not boost when backpropagating.
\item \textbf{Target network}: keep a separate DQN so that
one is used to compute the target values and the other one accumulates
the weight updates, which are periodically loaded onto the first one. This
avoid oscillations in the policy upon small changes to Q-values.
\end{itemize}
However, DQNs are meant for problems with a few possible actions,
and are therefore not appropriate for continuous space actions, like in
our case. Nevertheless, a recently proposed Deep RL algorithm referred to as
\textbf{Deep Deterministic Policy Gradient} or DDPG
(\citep{lillicrap2015continuous})
naturally accommodates this kind of problems.
It combines the actor-critic classical RL
approach \citep{sutton1998reinforcement} with Deterministic 
Policy Gradient \citep{silver2014deterministic}.
The original formulation of the policy gradient algorithm was proposed
in \citep{sutton1999policy}, which 
proved the policy gradient theorem
for a stochastic policy $\pi(s,a; \theta)$:

\begin{theorem} \label{theo:policygradient}
(\textbf{Policy Gradient theorem from \citep{sutton1999policy}})
For any MDP, if the parameters $\theta$ of the policy are updated proportionally
to the gradient of its performance $\rho$ 
then $\theta$ can be assured
to converge to a locally optimal policy in $\rho$, being the gradient
computed as
\[
\Delta \theta \approx \alpha \dfrac{\partial \rho}{\partial \theta} =
\alpha \sum_s d^{\pi}(s)\sum_a \dfrac{\partial \pi(s,a)}{\partial \theta}Q^\pi(s,a)
\]

with $\alpha$ being a positive step size and where
$d^\pi$ is defined as the discounted weighting of states encountered
starting at $s_0$ and then following $\pi$:
$d^\pi(s) = \sum_{t=0}^\infty \gamma^t P(s_t=s| s_0, \pi)$
\end{theorem}

This theorem was further extended in the same article
for the case where an approximation function $f$
is used in place of the policy $\pi$. In this conditions
the theorem holds valid as long as the weights of 
the approximation tend to zero upon convergence.
In our reference articles \citep{silver2014deterministic} and
\citep{lillicrap2015continuous}, the authors propose to use a
\textbf{deterministic policy} (as opposed to stochastic)
approximated by a neural network actor $\pi(s; \theta^\pi)$ that
depends on the state of the environment $s$ and has
weights $\theta^\pi$, and another separate network $Q(s,a; \theta^Q)$ implementing 
the critic, which is updated by means of the Bellman equation
like DQN \eqref{eq:dqngradient}:
\begin{equation}
Q(s_t, a_t) = \mathbb{E}_{r_t, s_{t+1}} \left[
r(s_t, a_t) + \gamma Q(s_{t+1}, \pi(s_{t+1}))
\right]
\end{equation}
And the actor is updated by applying the chain rule to the loss
function \eqref{eq:dqnloss} and updating the weights $\theta^\pi$ by following
the gradient of the loss with respect to them:
\begin{equation}
\begin{split}
\nabla_{\theta^\pi} \mathscr{L} & \approx
\mathbb{E}_s \left[ \nabla_{\theta^\pi} Q(s, \pi(s|\theta^\pi) | \theta^Q) \right]\\
& = \mathbb{E}_s \left[ \nabla_a Q(s, a | \theta^Q) |_{a=\pi(s|\theta^\pi)} \nabla_{\theta^\pi} \pi(s|\theta^\pi) \right]
\end{split}
\end{equation}
In order to introduce exploration behaviour, thanks to the DDPG
algorithm being off-policy, we can add random noise $\mathscr{N}$ to the policy.
This enables the algorithm to try unexplored areas from the action
space to discover improvement opportunities, much like the role of $\varepsilon$
in $\varepsilon$-greedy policies in Q-learning.\\

In order to improve stability, DDPG also can be applied the same measures
as DQNs, namely reward clipping, experience replay (by means of a
\textit{replay buffer} referred to as $R$ in algorithm \ref{alg:ddpg}) and separate
target network. In order to implement this last measure for DDPG,
two extra target actor and critic networks (referred to as $\pi'$
and $Q'$ in algorithm \ref{alg:ddpg}) to compute the target Q values, separated
from the normal actor and critic (referred to as $\pi$ and $Q$ in
algorithm \ref{alg:ddpg}) that are updated at every step and which weights
are used to compute \textit{small updates} to the target networks.
The complete DDPG, as proposed in
\citep{lillicrap2015continuous}, is summarized in algorithm
\ref{alg:ddpg}.

\begin{algorithm}[h]
\caption{Deep Deterministic Policy Gradient algorithm} \label{alg:ddpg}
\begin{algorithmic}
\State Randomly initialize weights of $Q(s,a|\theta^Q)$ and
       $\pi(s|\theta^\pi)$.
\State Initialize target net weights
       $\theta^{Q'} \leftarrow \theta^Q$, $\theta^{\pi'} \leftarrow \theta^\pi$.
\State Initialize replay buffer $R$.
\For{each episode}
  \State Initialize random process $\mathscr{N}$ for action exploration.
  \State Receive initial observation state $s_1$.
  \For{each step $t$ of episode}
    \State Select action $a_t = \pi(s_t|\theta^\pi) + \mathscr{N}_t$.
    \State Execute $a_t$ and observe reward $r_t$ and state $s_{t+1}$.
    \State Store transition $(s_t, a_t, r_t, s_{t+1})$ in $R$.
    \State Sample from $R$ a minibatch of $N$ transitions.
    \State Set $y_i= r_i + \gamma Q'(s_{i+1},\pi'(s_i+1|\theta^{\pi'})|\theta^{Q'})$.
    \State Update critic by minimizing the loss:
    \State \hspace{5mm} $\mathscr{L}=\dfrac{1}{N}\sum_i (y_i - Q(s_i,a_i|\theta^Q))^2$.
    \State Update the actor using the sampled policy gradient:
    \State \hspace{5mm} \mbox{$\nabla_{\theta^\pi} \mathscr{L} \approx \dfrac{1}{N} \sum_i
            \nabla_a Q(s,a|\theta^Q)
             \nabla_{\theta^\pi} \pi(s|\theta^\pi)|s_i$}.
    \State Update the target networks:
    \State \hspace{5mm} $\theta^{Q'} \leftarrow \tau \theta^Q + (1-\tau) \theta^{Q'}$
    \State \hspace{5mm} $\theta^{\pi'} \leftarrow \tau \theta^\pi+ (1-\tau) \theta^{\pi'}$.
  \EndFor
\EndFor
\end{algorithmic}
\end{algorithm}

\section{Proposed Approach} \label{chap:approach}

In this section we explain the approach we are proposing to address the control of
urban traffic lights, along with the
rationale that led to it.
We begin with section
\ref{sec:inputinfo} by defining which information shall be used
as input to our algorithm among all the data that is available from our simulation
environment. We proceed by choosing a problem representation for such information to
be fed into our algorithm in section \ref{sec:inputrepresentation} for the traffic
state and section \ref{sec:rewards} for the rewards.

\subsection{Input Information} \label{sec:inputinfo}

The fact that we are using a simulator to evaluate
the performance of our proposed
application of deep learning to traffic control, makes
the traffic state fully observable
to us.  However, in order for our system to be applied
to the real world, it must be possible for
our input information to be derived from data that
is available in a typical urban traffic setup.
The most remarkable examples of readily available
data are the ones sourced by \textbf{traffic detectors}.
They are sensors located throughout the traffic network
that provide measurements about the
traffic passing through them. Although there are different
types of traffic detectors, the most
usual ones are induction loops placed under the pavement
that send real time information about
the vehicles going over them. The information
that can normally be taken from such type of detectors
comprise vehicle count (number of vehicles that went
over the detector during the sampling period), 
vehicle average speed during the sampling period and
occupancy (the percentage of time in which there was a vehicle located over
the detector).
This way, we decide to constrain the information received about
the state of the network to vehicle counts, average speed and occupancy of every
detector in our traffic networks, along with the description of the network
itself, comprising the location of all roads, their connections, etc.

\subsection{Congestion Measurement} \label{sec:speedscore}

Following the self-imposed constraint to use only data that is actually available
in a real scenario, we shall elaborate a summary of the state of the
traffic based on vehicle counts, average speeds and occupancy.
This way, we defined a measured called
\textbf{speed score}, that is defined for detector $i$ as: 
\begin{equation} \label{eq:speedscore}
\mathit{speed\_score}_i = 
\min \left( \dfrac{\mathit{avg\_speed}_i}{\mathit{max\_speed}_i} , 1.0\right)
\end{equation}
where $\mathit{avg\_speed}_i$ refers to the average of the speeds
measured by traffic detector $i$ and $\mathit{max\_speed}_i$
refers to the maximum speed in the road where detector $i$ is
located.
Note that the speed score hence ranges in $[0, 1]$.
This measure will be the base to elaborate the
representation of both the state of the environment
(section \ref{sec:inputrepresentation}) and 
the rewards for our reinforcement learning algorithm
(section \ref{sec:rewards}).

\subsection{Data Aggregation Period}

The microscopic traffic simulator used for our
experiments divides the
simulation into steps. At each step, a small fixed amount
of time is simulated and the state of the vehicles
(e.g. position, speed, acceleration) is
updated according to the dynamics of the system. 
This amount of time is configured to be 0.75 seconds by
default, and we have kept this parameter.
However, such an amount of time is too short to imply
a change in the vehicle counts of the detectors. Therefore,
it is needed to have a larger period over which the data
is aggregated; we refer to this period
as \textbf{\textit{episode step}}, or simply "step" when
there is no risk of confusion.
This way, the data is collected at each
simulation step and then it is aggregated every
episode step for the DDPG algorithm to receive
it as input. In order to properly combine the speed scores
of several simulation steps, we take their weighted
average, using the proportion of vehicle counts.
In an analogous way, the traffic light timings generated by the DDPG
algorithm are used during the following episode step.
The duration of the episode step was chosen by means
of grid search, determining an optimum value of 120 seconds.

\subsection{State Space} \label{sec:inputrepresentation}

In order to keep a state vector of the environment, we make direct
use of the speed
score described in section \ref{sec:speedscore}, as it not only summarizes
properly the congestion of the network, but also incorporates the notion
of maximum speed of each road.
This way, the state vector has one component per detector, each one
defined as shown in \eqref{eq:statedef}.
\begin{equation} \label{eq:statedef}
\mathit{state}_i = \mathit{speed\_score}_i
\end{equation}
The rationale for choosing the speed score is that, 
the higher the speed score, the higher the speed of the
vehicles relative to the maximum speed of the road, and
hence the higher the traffic flow.

\subsection{Action Space} \label{sec:actions}

In the real world there are several instruments to dynamically regulate
traffic: traffic lights, police agents, traffic information displays,
temporal traffic signs (e.g. to block a road where there is an accident),
etc. Although it is possible to apply
many of these alternatives in traffic simulation software, we opted
to keep the problem at a manageable level and constrain the actions
to be applied only to traffic lights.
The naive approach would be to let our agent simply control
the traffic lights directly by setting the color of each traffic light
individually at every simulation step, that is, the actions generated by our agent
would be a list with the color (red, green or yellow)
for each traffic light.  
However, traffic lights in an intersection are synchronized:
when one of the traffic lights of the intersection is green,
the traffic in the perpendicular direction is forbidden
by setting the traffic lights of such a direction to red.
This allows to \textit{multiplex} the usage of the intersection.
Therefore, letting our agent freely control the colors of the
traffic lights would probably lead to chaotic situations.
In order to avoid that, we should keep the \textit{phases} of the
traffic lights in each intersection. With that premise, 
we shall only control the \textit{phase duration}, hence the dynamics are kept
the same, only being accelerated or decelerated.
This way, if the network
has $N$ different phases, the action vector has $N$ components, each
of them being a real number that has a
scaling effect on the duration of the phase.
However, for each intersection, the total duration of the cycle
(i.e. the sum of all phases in the intersection) should be kept unchanged.
This is important because in most cases, the cycles of nearby
intersections are synchronized so that vehicles travelling from one intersection
to the other can catch the proper phase, thus improving the traffic flow.
In order to ensure
that the intersection cycle is kept, the scaling factor of the
phases from the same intersection are passed through a softmax
function (also known as normalized exponential function). The result
is the ratio of the phase duration over the total cycle duration.
In order to ensure a minimum phase duration, the scaling factor
is only applied to 80\% of the duration.

\subsection{Rewards} \label{sec:rewards}

The role of the rewards is to provide feedback to the reinforcement learning  algorithm
about the performance of the actions taken previously.
As commented in previous section, it would be possible for us to define a
reward scheme that makes use of
information about the travel times of the vehicles.
However, as we are self-constraining to the information that is available
in real world scenarios, we can not rely on other measures apart from
detector data, e.g. vehicle counts, speeds. This way, we shall use the
speed score described in section \ref{sec:speedscore}. But the speed score
alone does not tell whether the actions taken by our agent actually
improve the situation or make it worse. Therefore, in order to capture
such information, we shall introduce the concept of \textbf{\textit{baseline}},
defined as the speed score for a detector during a hypothetical simulation that is
exactly like the one under evaluation but with no intervention by
the agent, recorded at the same time step.
This way, our reward is the difference between the speed score and
the baseline, scaled by the
vehicle counts passing through each detector (in order to give more
weight to scores where the number of vehicles is higher), and further
scaled by a factor $\alpha$ to keep the reward in a narrow range,
as shown in \eqref{eq:reward}.
\begin{equation} \label{eq:reward}
\mathit{reward}_i =
\alpha \cdot \mathit{count}_i \cdot
\left( \mathit{speed\_score}_i 
- \mathit{baseline}_i \right)
\end{equation}
Note that we may want to normalize the weights by dividing by the total
vehicles traversing all the detectors. This would restrain the rewards in
the range $[-1, +1]$. This, however, would make the rewards obtained in
different simulation steps not comparable (i.e. a lower total number of
vehicles in the simulation at instant $t$ would lead to higher rewards).
The factor $\alpha$ was chosen to be $1/50$ empirically, by observing
the unscaled values of different networks and choosing a value in an
order of magnitude that leaves the scaled value around $1.0$. This is
important in order to control the scale of the resulting gradients.
Another alternative used in \citep{mnih2013playing, mnih2015human} with
this very purpose is reward clipping; this, however, implies losing
information about the scale of the rewards. Therefore, we chose to
apply a proper scaling instead.
There is a reward computed for each detector at each simulation time
step. Such rewards are not combined in any way, but are all used for the
DDPG optimization, as described in section \ref{sec:multirewards}.
Given the stochastic nature of the microsimulator used,
the results obtained depend on the random seed set for
the simulation. This way, when computing the reward, the baseline is
taken from a simulation with the same seed as the one under evaluation.

\subsection{Deep Network Architecture}

\begin{figure}[h] 
\centering
\vspace{-2.8mm}
\includegraphics[width=.5\linewidth]{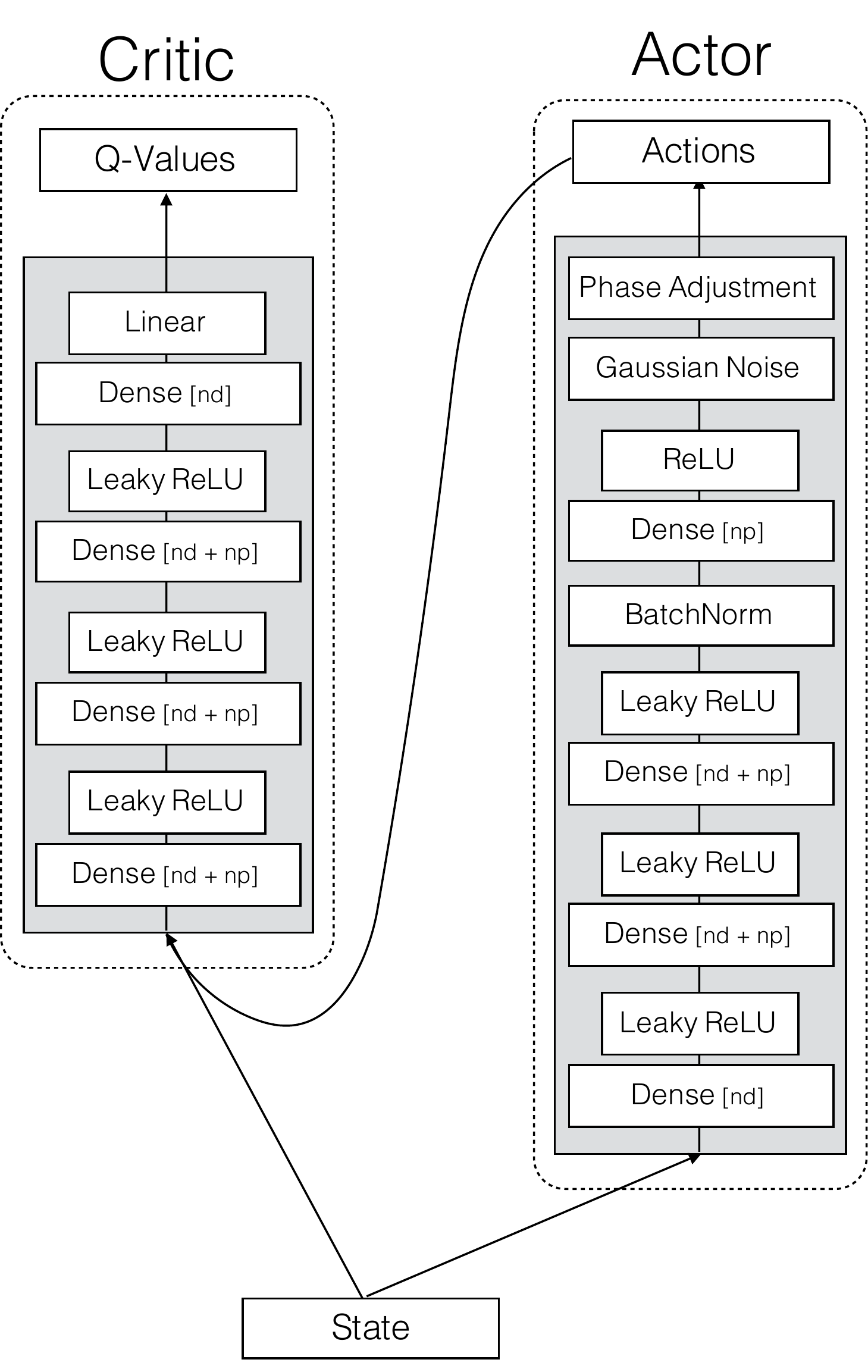}
\caption{Critic and Actor networks}
\label{fig:basicarchitecture}
\end{figure}

Our neural architecture consists in a Deep Deterministic Actor-Critic
Policy Gradient approach. 
It is comprised of two networks: the actor network $\pi$
and the critic network $Q$.
The \textbf{actor network}
receives the current state of the simulation
(as described in section \ref{sec:inputrepresentation})
and outputs the actions, as described in \ref{sec:actions}.
As shown in figure \ref{fig:basicarchitecture}, the network is
comprised of several layers. It starts with several fully connected
layers (also known as \textit{dense} layers) with Leaky ReLU
activations \citep{maas2013rectifier}, where the number of units
is indicated in brackets, with $\mathit{nd}$ being the number
of detectors in the traffic network and $\mathit{np}$ is the
number of phases of the traffic network.
Across those many layers, the width of the network increases
and then decreases, up to having as many units as actions, that
is, the last mentioned dense layer has as many units as traffic light
phases in the network.
At that point, we introduce a batch normalization layer and another
fully connected layer with ReLU activation.
The output of the last mentioned layer are real numbers in
the range $[0, +\infty]$, so we should apply some kind of
transformation that allows us to use them as scaling factors
for the phase durations (e.g. clipping to the range $[0.2, 3.0]$).
However, as mentioned in section \ref{sec:actions},
we want to keep the traffic light cycles constant. Therefore,
we shall apply an element-wise scaling computed on the summation
of the actions of the phases in the same traffic light cycle,
that is, for each \textit{scaling factory} we divide by the sum
of all the factors for phases belonging to the same group
(hence obtaining the new ratios of each phase over the cycle
duration) and then multiply by the original duration of the cycle.
In order to keep a minimum duration for each phase, such
computation is only applied to the 80\% of the duration
of the cycle. Such a computation can be pre-calculated into
a matrix, which we call the \textit{phase adjustment matrix}, which
is applied in the layer labeled as "Phase adjustment" in figure
\ref{fig:basicarchitecture}, and which finally gives the scaling
factors to be applied to phase durations. This careful scaling meant
to keep the total cycle duration can be ruined by the exploration
component of the algorithm, as described in \ref{alg:ddpg}, which
consists of adding noise to the actions (and therefore likely breaking the
total cycle duration), This way, we implement the injection of noise
as another layer prior to the phase adjustment.
The \textbf{critic network} receives the current state of the simulation
plus the action generated by the actor, and outputs the Q-values
associated to them. Like the actor, it is comprised of several
fully connected layers with leaky ReLU activations, plus a final
dense layer with linear activation.

\subsection{Disaggregated Rewards} \label{sec:multirewards}

In our reference article \citep{lillicrap2015continuous}, as well as all
landmark ones like \citep{mnih2013playing} and \citep{mnih2015human},
the reward is a single scalar value. However, in our case we build
a reward value for each detector in the network.
One option to use such a vector of rewards could be to \textit{scalarize}
them into a single value. This, however, would imply losing valuable
information regarding the location of the effects of the actions taken
by the actor. Instead, we will keep then disaggregated,
leveraging the structure of the DDPG algorithm, which
climbs in the direction of the gradient of the critic. This is partially
analogous to a regression problem on the Q-value and hence does
not impose constraints on the dimensionality of the rewards.
This way, we will have a $N$-dimensional reward vector, where $N$
is the number of detectors in the network.
This extends the policy gradient theorem from \citep{silver2014deterministic}
so that the reward function is no longer
defined as $r: S \times A \rightarrow \mathbb{R}$
but as $r: S \times A \rightarrow \mathbb{R}^N$. This is analogous to
having $N$ agents sharing the same actor and critic networks
(i.e. sharing weights $\theta^\pi$ and $\theta^Q$) and being trained
simultaneously over $N$ different unidimensional reward functions.
This, effectively, implements multiobjective reinforcement learning.
To the best of our knowledge,
the use of \textbf{\textit{disaggregated rewards}}
has not been used before in the reinforcement learning literature.
Despite having proved useful in our experiments, further study is needed
in order to fully characterize the effect of disaggregated rewards
on benchmark problems. This is one of the future lines of research that
can be spawned from this work.
Such an approach could be further
refined by weighting rewards according to traffic control expert
knowledge, which will then be incorporated in the computation of the
policy gradients.

\subsection{Convergence} \label{sec:gammaschedule}

There are different aspects that needed to be properly tuned
in order for the learning to achieve convergence:

\begin{itemize}
\item{Weight Initialization} has been a key issue in the results cast by deep
learning algorithms. The early architectures could only achieve acceptable
results if they were pre-trained by means of unsupervised learning so that
they could have \textit{learned} the input data structure \citep{erhan2010does}.
The use of sigmoid or hyperbolic
tangent activations makes it difficult to optimize neural networks 
due to the numerous local
minima in the function loss defined over the parameter space. 
With pre-training, the exploration of the parameter space does not
begin in a random point, but in a point that \textit{hopefully} is not
too far from a good local minimum.
Pretraining became no longer necessary to achieve convergence
thanks to the use of rectified linear activation units (ReLUs)
\citep{nair2010rectified} and sensible
weight initialization strategies.
In our case, different random weight initializations (i.e. Glorot's
\citep{glorot2010understanding} and He's \citep{he2015delving}) gave the best
results, finally selecting He's approach.

\item{Updates to the Critic}:
after our first experiments it became evident the divergence of the
learning of the network. Careful inspection of the algorithm
byproducts revealed that the cause of the divergence was 
that the critic network $Q'$ predicted higher outcomes at every iteration,
as trained according to equation \eqref{eq:closedloop} extracted
from algorithm \ref{alg:ddpg}.
\begin{equation} \label{eq:closedloop}
y_i= r_i + \gamma Q'(s_{i+1},\pi'(s_i+1|\theta^{\pi'})|\theta^{Q'})
\end{equation}
As DDPG learning -like any other reinforcement learning with value function
approximation approach- is a closed
loop system in which the target value at step $t+1$
is biased by the training at steps $t$, drifts 
can be amplified, thus ruining the learning, as the distance between the
desired value for $Q$ and the obtained one differ more
and more.
In order to mitigate this divergence problem, our
proposal consists in reducing the coupling by means
of the application of a schedule on the value of the
discount factor $\gamma$ from Bellman's equation,
which is shown in figure \ref{fig:gamma}.
\begin{figure}[h] 
\centering
\includegraphics[width=.8\linewidth]{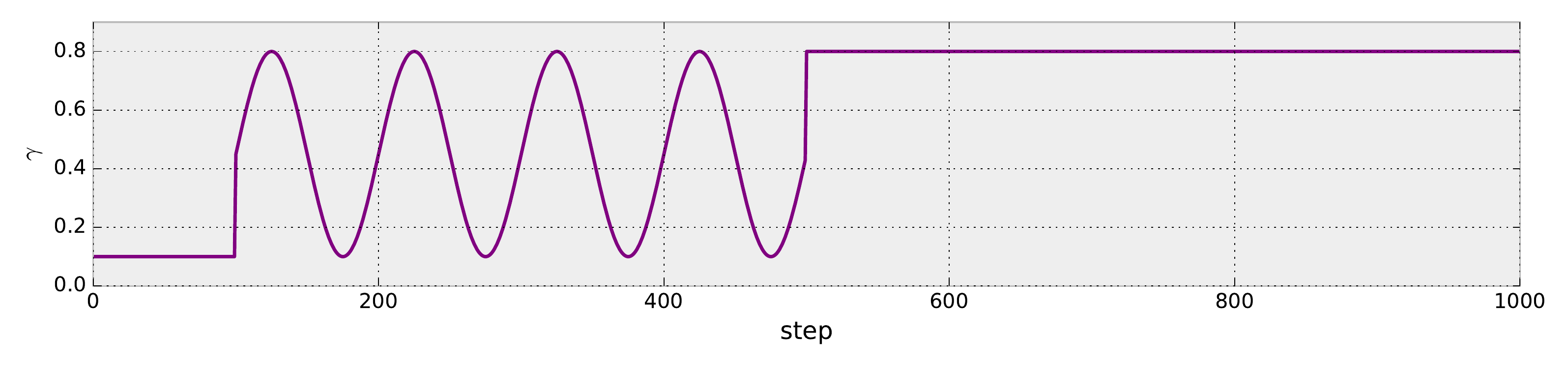}
\caption{Schedule for the discount factor $\gamma$.}
\label{fig:gamma}
\end{figure}
The schedule of $\gamma$ is applied at the level of the experiment,
not within the episode. The oscillation in $\gamma$ shown in figure
\ref{fig:gamma} is meant to enable the critic network not to enter
in the regime where the feedback leads to divergence.
Discount Factor
scheduling has been proposed before in \citep{harrington2013robot}
with positive results, although in that case the schedule consisted
in a decaying rate.

\item{Gradient evolution}: the convergence of
the algorithm can be evaluated thanks to
the norm of the gradient used to update the actor network $\pi$.
If such a norm decreases over time and stagnates around a low
value, it is a sign that the algorithm has reached a stable point
and that the results might not further improve. This way, in
the experiments described in subsequent sections, monitoring of
the gradient norm is used to track progress.
The gradient norm can also be controlled in order to avoid too
large updates that make the algorithm diverge, e.g.
\citep{mnih2013playing}. This mechanism
is called \textit{gradient norm clipping} and consists of scaling
the gradient so that its norm is not over a certain value. Such a
value was empirically established as $0.5$ in our case.

\end{itemize}

\subsection{Summary}

Our proposal is to apply Deep Deterministic Policy
Gradient, as formulated in \citep{lillicrap2015continuous}, to
the traffic optimization problem by controlling the traffic
lights timing. We make use of a multilayer perceptron type
of architecture, both for the actor and the critic networks.
The actor is designed so that the modifications to the traffic
light timings keep the cycle duration. In order to optimize
the networks we make use of stochastic gradient descent. In
order to improve convergence, we make use of a replay memory,
gradient norm clipping and a schedule for the discount
rate $\gamma$. The input state used to feed
the network consists of traffic detector information, namely
vehicle counts and average speeds, which are combined in a
single \textit{speed score}. The rewards used as reinforcement
signal are the improvements over the measurements without
any control action being performed (i.e. baseline). Such rewards
are not aggregated but fed directly as expected values of 
the critic network.

\section{Experiments} \label{chap:experiments}

In this section we describe the experiments conducted in order to evaluate
the performance of the proposed approach. In
section \ref{sec:experimentdesign} we show the different traffic scenarios used
while in section \ref{sec:experimentresults} we describe the results obtained
in each one, along with lessons learned from the problems found, plus hints
for future research.

\subsection{Design of the Experiments} \label{sec:experimentdesign}

In order to evaluate our deep RL algorithm, we devised increasing
complexity traffic networks. 
For each one, we applied our \textbf{DDPG} algorithm to control the traffic
light timing, but also applied \textbf{multi-agent Q-Learning} and
\textbf{random timing} in order to have a reference to properly assess
the performance of our approach.
At each experiment, the \textbf{DDPG} algorithm receives as input the 
information of all detectors in the network, and generates the timings
of all traffic light phases.
In the multi-agent \textbf{Q-learning} implementation,
there is one agent managing each intersection phase. It receives the
information from the closest few detectors and generates the timings
for the aforementioned phase. Given the tabular nature of Q-learning,
both the state space and the action space need to be categorical.
For this, tile coding is used. Regarding the state space, the tiles
are defined based on the same state space values as DDPG (see section
\ref{sec:inputrepresentation}), clustered in one the following
4 ranges
$[-1.0, -0.2]$, $[-0.2, -0.001]$, $[-0.001, 0.02]$, $[0.02, 1.0]$,
which were chosen empirically. As one Q-learning agent controls
the $N_i$ phases of the traffic lights of an intersection $i$, the
number of states for an agent is $4^{N_i}$. The action space
is analogous, being the generated timings one of the values
$0.2$, $0.5$, $1.0$, $2.0$ or $3.5$. The selected ratio
(i.e. ratio over the original phase duration) is applied to
the duration of the phase controlled by the Q-learning agent.
As there is one agent per phase, this is a multi-agent reinforcement
learning setup, where agents do not communicate with each other.
They do have overlapping inputs, though, as the data from a
detector can be fed to the agents of several phases. In order to
keep the cycle times constant, we apply the same phase adjustment
used for the DDPG agent, described in section
\ref{sec:actions}.
The \textbf{random} agent generates random timings in the range
$[0, 1]$, and then the previously mentioned phase adjustment is applied to keep
the cycle durations constant (see section \ref{sec:actions}).\\

Given the stochastic nature of the microscopic traffic
simulator used, the results obtained at the
experiments depend on the random seed set for
the simulation. In order to address the implications of this,
we do as follows:
\begin{itemize}
\item In order for the algorithms not to overfit to the dynamics
of a single simulation, we randomize the seed of each simulation.
We take into account this also for the computation of the
baseline, as described in section \ref{sec:rewards}.
\item We repeat the experiments several times, and present the
results over all of them (showing the average, maximum or
minimum data depending on the case).
\end{itemize}

\subsection{Network A} \label{sec:simplea}

\begin{figure}[H] 
\centering    
\includegraphics[width=.42\linewidth]{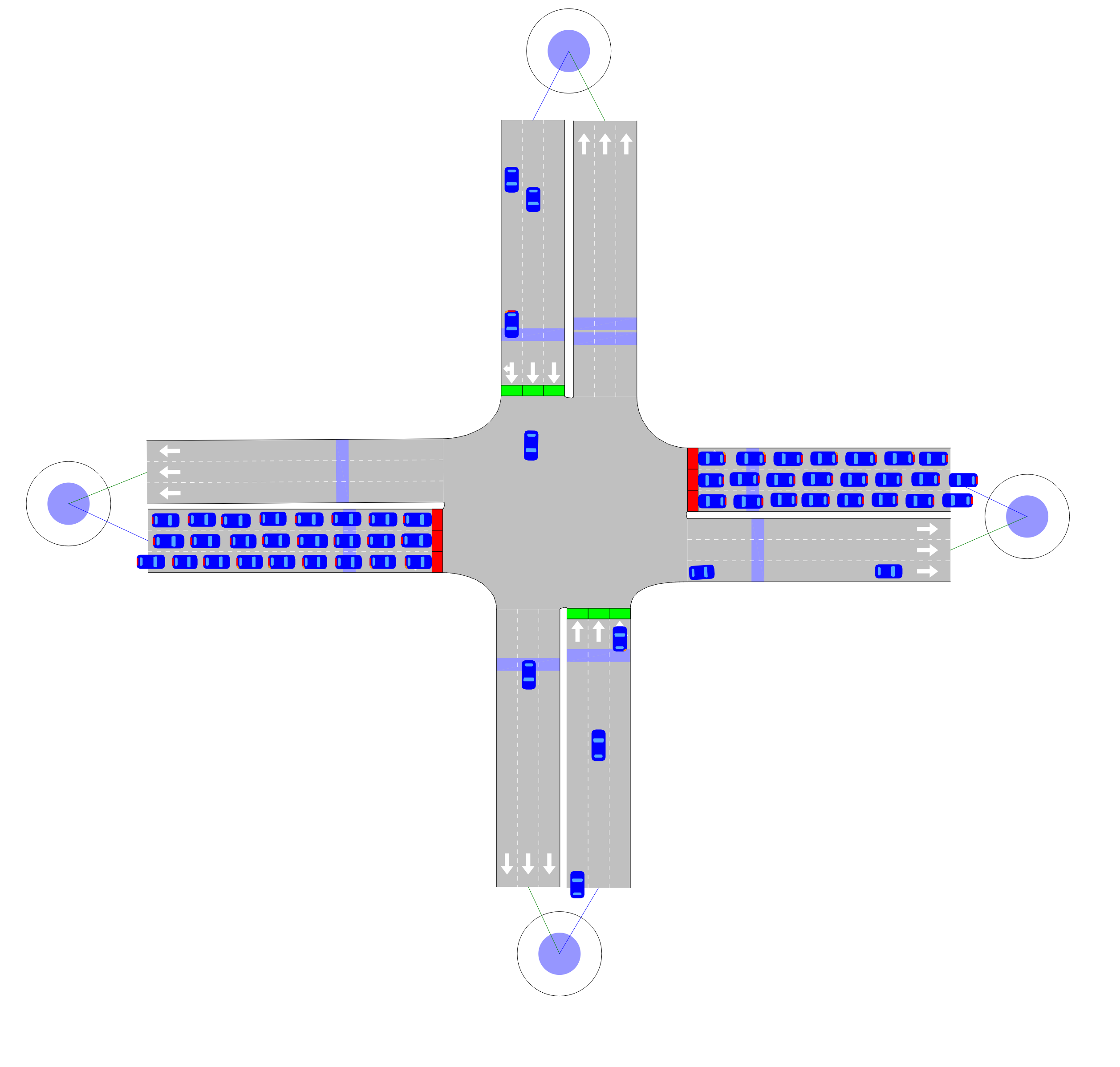}
\caption{Urban network A}
\label{fig:neta}
\end{figure}

This network, shown in figure \ref{fig:neta} consists only
of an intersection of two 2-lane roads. At the
intersection vehicles can either go straight or turn to their right. It is
forbidden to turn left, therefore simplifying the traffic dynamics and
traffic light phases. There are 8 detectors (in each road there is
one detector before the intersection and another one after it).
There are two phases in the traffic light group: phase 1 allows
horizontal traffic while phase 2 allows vertical circulation. Phase 1
lasts 15 seconds and phase 2 lasts 70 seconds, with a 5-seconds
inter-phase. Phases 1 and 2 have unbalanced duration on purpose, to
have the horizontal road accumulate vehicles for long time. This
gives our algorithm room to easily improve the traffic flow with
phase duration changes.
The simulation comprises 1 hour and the vehicle demand
is constant: for each pair of centroids, there are 150 vehicles.

The traffic demand is defined by hand, with the proper order of
magnitude to ensure congestion. The definition and duration of the
phases were computed by means of the classical cycle length
determination and green time allocation formulas from
\citep{webster1958traffic}.

\subsection{Network B}

\begin{figure}[H] 
\centering    
\includegraphics[width=1.0\linewidth]{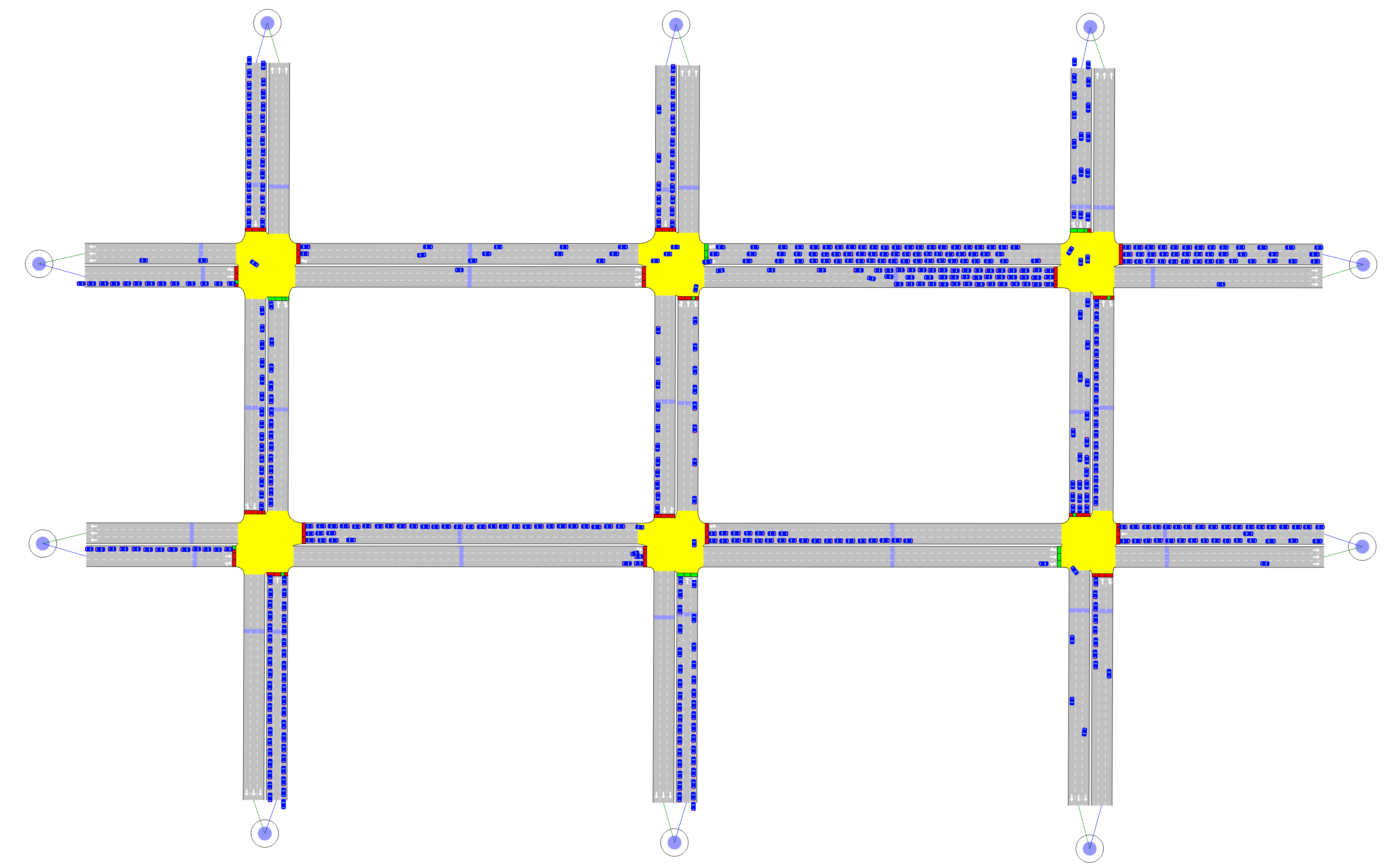}
\caption{Urban network B}
\label{fig:netb}
\end{figure}

This network, shown in figure \ref{fig:netb} consists of a grid
layout of 3 vertical roads and 2 horizontal ones, crossing in 6
intersections that all have traffic lights.
Traffic in an intersection
can either go straight, left or right, that is, all turns are allowed,
complicating the traffic light phases, which have been generated
algorithmically by the software with the optimal timing, totalling
30 phases. There are detectors before and after each intersection, totalling 
17 detectors. The traffic demand is defined by hand, ensuring
congestion. The trafic light phases were defined, like network A,
with the classical approach from \citep{webster1958traffic}.
Four out of six junctions have 5 phases,
while the remaining two junctions have 4 and 6 phases each.
The traffic demand has been created in a random manner, but ensuring enough vehicles
are present and trying to collapse some of the sections of the network.

\subsection{Network C}

\begin{figure}[H] 
\centering    
\includegraphics[width=1.\linewidth]{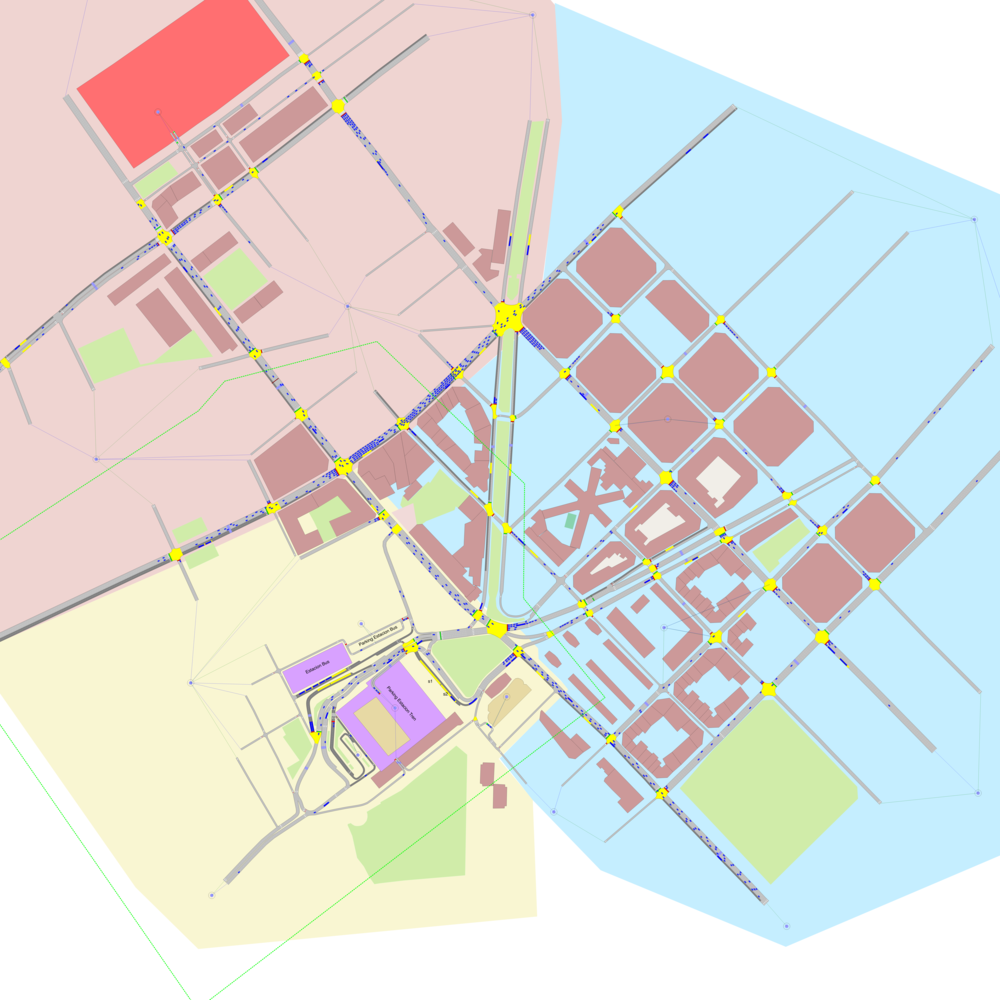}
\caption{Urban network C}
\label{fig:netc}
\end{figure}

This network, shown in figure \ref{fig:netc} is a replica of the Sants
area in the city of Barcelona (Spain). 
There are 43 junctions, totalling 102 traffic light phases,
and 29 traffic detectors. 
The locations of the detectors matches the real world.
The traffic demand matches that of the peak hour in Barcelona, and it presents
high degree of congestion.
The number of controlled
phases per junction \footnote{Note that phases from the network that
have a very small duration (i.e. 2 seconds or less) are excluded from
the control of the agent} ranges from 1 to 6, having most of them
only two phases.

%
\subsection{Results} \label{sec:experimentresults}

In order to evaluate the performance of our DDPG approach
compared to both normal Q-learning and random timings on
each of our test networks, our main reference measure
shall be the episode average reward (note that, as described
in section \ref{sec:rewards} there is actually a vector of rewards,
with one element per detector in the network, that is why we
compute the average reward) of the \textit{best}
experiment trial, understanding "best" experiment as the one
where the maximum episode average reward was obtained.
\begin{figure}[H] 
\centering    
\includegraphics[width=.8\linewidth]{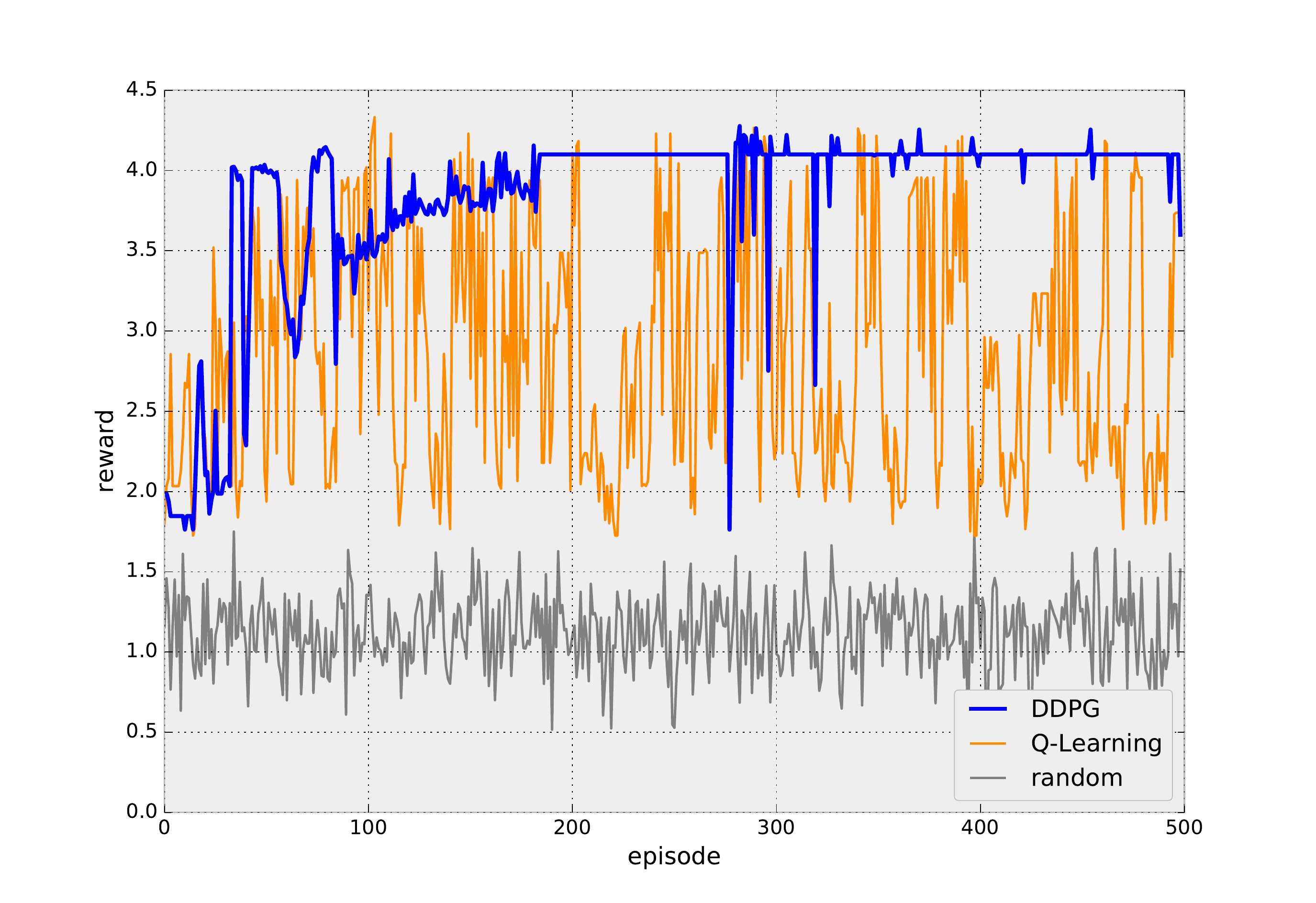}
\caption{Algorithm performance comparison on network A}
\label{fig:performancea}
\end{figure}
In figure \ref{fig:performancea} we can find the performance
comparison for network A. Both the DDPG approach and the
classical Q-learning reach the same levels of reward.
On the other hand, it is noticeable the differences in
the convergence of both approaches: while Q-learning is
unstable, DDPG remains remarkably stable once it reached
its peak performance.
\begin{figure}[H] 
\centering    
\includegraphics[width=.75\linewidth]{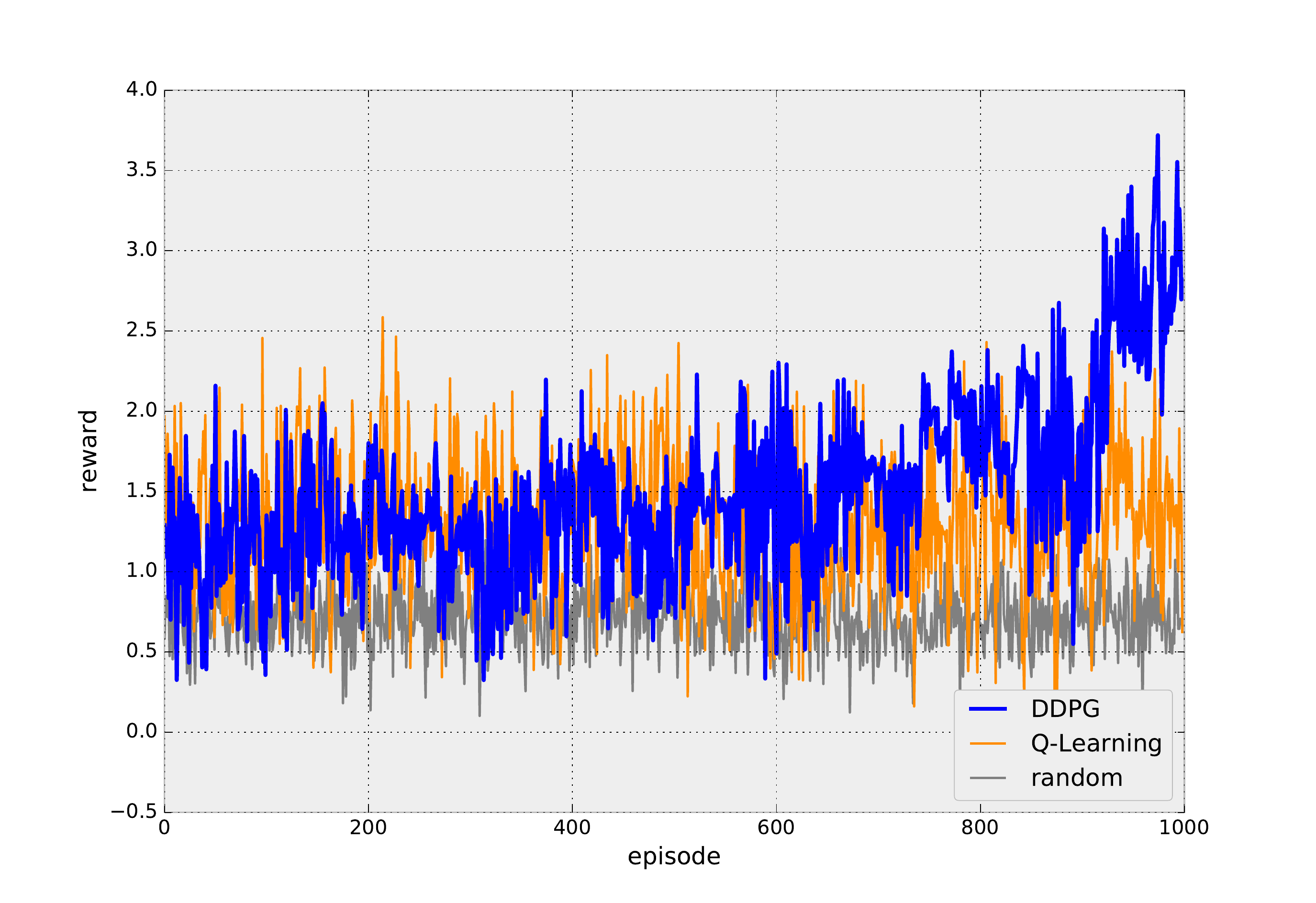}
\caption{Algorithm performance comparison on network B}
\label{fig:performanceb}
\end{figure}
In figure \ref{fig:performanceb} we can find the performance
comparison for network B. While Q-learning maintains the same
band of variations along the simulations, DDPG starts to converge.
Given the great computational costs of running the full set of
simulations for one network, it was not affordable to
let it run indefinitely, despite the promising trend.
\begin{figure}[H] 
\centering    
\includegraphics[width=.75\linewidth]{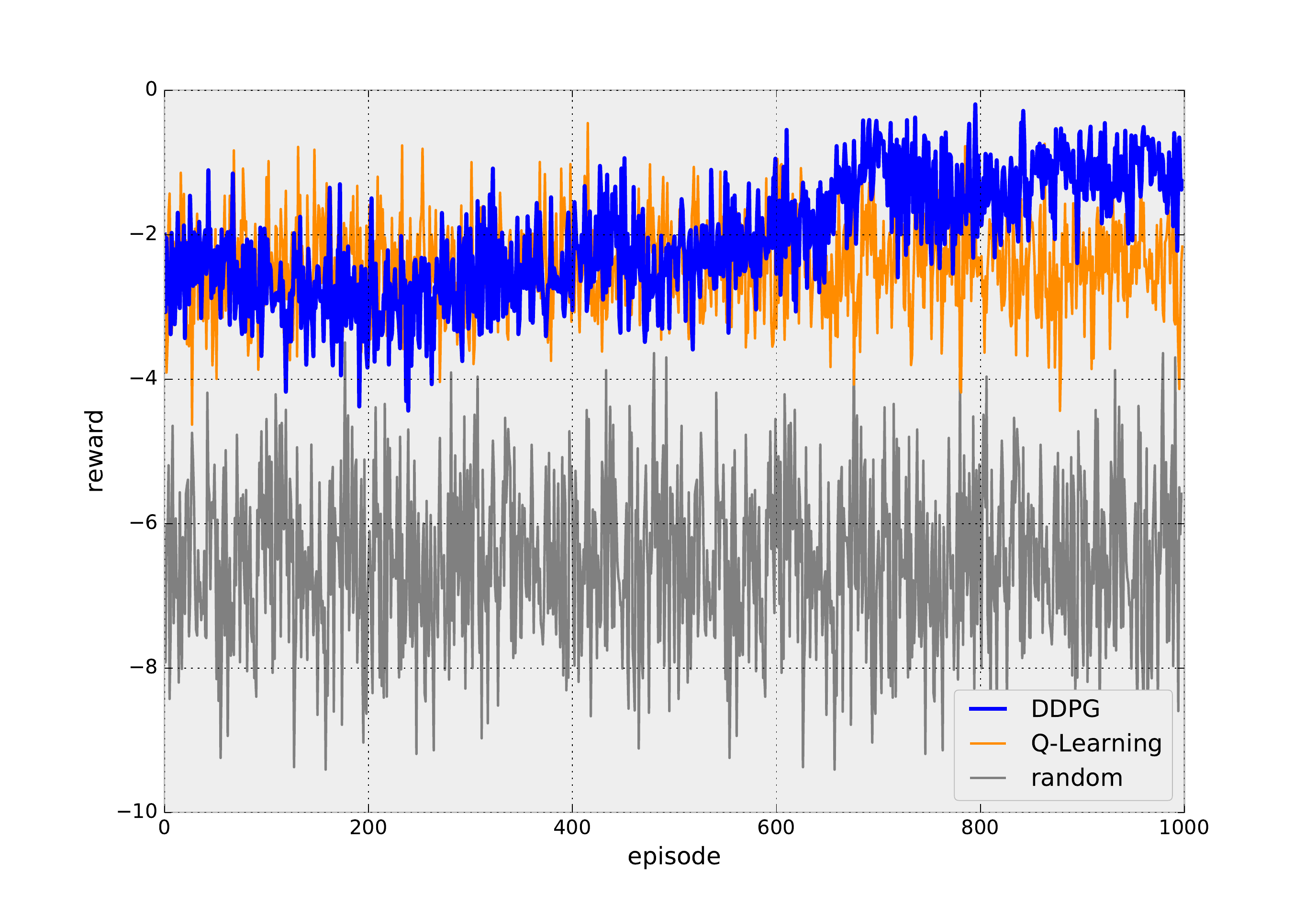}
\caption{Algorithm performance comparison on network C}
\label{fig:performancec}
\end{figure}
Figure \ref{fig:performancec} shows the performance
comparison for network C, from which we can appreciate
that both DDPG and Q-learning performs at the same level,
and that such a level is beneath zero, from which we know that
they are actually worse than doing nothing. 
This way, the performance of DDPG is clearly superior to Q-learning for
the simplest scenario (network A), slightly better for scenarios with a
few intersections (network B) and at the same level for real world networks.
From the evolution of the gradient for medium and large networks,
we observed
that convergence was not achieved, as it remains always at the maximum
value induced by the gradient norm clipping. This suggests that the
algorithm needs more training time to converge (probable for network B)
or that it diverges (probable for network C). In any case, further study
would be needed in order to assess the needed training times and the
needed convergence improvement techniques.

\section{Conclusions} \label{chap:conclusion}

We studied the application of Deep Deterministic
Policy Gradient (DDPG) to increasingly complex
scenarios. We obtained good results in network
A, which is analogous to most of the scenarios
used to test reinforcement learning applied to
traffic light control (see section \ref{chap:related}
for details on this); nevertheless, for such
a small network, vanilla Q-learning performs
\textit{on par}, but with less stability, though.
However, when the complexity of the network
increases, Q-learning can no longer scale,
while DDPG still can improve consistently the
obtained rewards. With a real world scenario,
our DDPG approach is not able to properly control
the traffic better than doing nothing. The
good trend for network B shown in figure
\ref{fig:performanceb}, suggests that longer
training time may lead to better results. This
might be also true for network C, but the
extremely high computational costs could not
be handled without large scale hardware
infrastructure.
Our results show that DDPG is able to better
scale to larger networks than classical tabular
approaches like Q-learning.
Therefore, DDPG is able
to address the \textit{curse of dimensionality}
\citep{Goodfellow-et-al-2016} regarding the traffic
light control domain, at least partially.
However, it is
not clear that the chosen reward scheme
(described in section \ref{sec:rewards}) is
appropriate. One of its many weaknesses is
its \textit{fairness} for judging the performance
of the algorithm based on the individual
detector information. In real life traffic
optimization it is common to favour some
areas so that traffic flow in arterials or
large roads is improved, at the cost of
worsening side small roads. The same principle
could be applied to engineer a more realistic
reward function from the point of view of
traffic control theory.\\

In order to properly asses the applicability of
the proposed approach to real world setups, it
would also be needed to provide a wide degree of
variations in the conditions of the simulation,
from changes in the traffic demand to having
road incidents in the simulation.
Another aspect that needs further study is the
effect of the amount and location of traffic
detectors on the performance of the algorithm.
In our networks A and B, there were detectors
at every section of the network, while in network
C their placement was scattered, which is the norm
in real world scenarios. We appreciate a loose
relation between the \textit{degree of observability}
of the state of the network and the performance
of our proposed traffic light timing control
algorithm.
Further assessment about
the influence of observability
of the state of the network would help characterize
the performance of the DDPG algorithm and even
turn it into a means for choosing potential locations
for new detector in the real world. Also,
the \textit{relevance} of the provided information
is not the same for all detectors; some of them
may provide almost irrelevant information while
others are key for understanding the traffic state.
This is another aspect that should be further
studied, along with the effect of the noise
present in data delivered by real traffic detectors.
An issue regarding the performance of our 
approach is the sudden drops in the rewards
obtained through the training process. This
suggests that the landscape of the reward
function with respect to the actor and critic
network parameters is very irregular, which
leads the optimization to fall into
\textit{bad regions} when climbing in the
direction of the gradient. A possible future line of
research that addressed this problem could
be applying Trusted Region Policy Optimization \citep{schulman2015trust},
that is, leveraging the simulated nature of
our setup to explore more efficiently the
solution space. This would allow it to be
more data efficient, achieving comparable results
with less training.\\

We have introduced a
concept that, to the best of our
knowledge, has not been used before in the deep
reinforcement learning literature,
namely the use of disaggregated rewards (described
in section \ref{sec:multirewards}). This technique
needs to be studied in isolation from other factors
on benchmark problems
in order to properly assess its effect and contribution
to the performance of the algorithms. This is another possible
line of research to be spawned from this work.\\


On the other hand, we have failed to profit from the
geometric information about the traffic network.
This
is clearly a possible future line of research, that
can leverage recent advances in the application of
convolutional networks to arbitrary graphs, similar
to \citep{defferrard2016convolutional}.\\


Finally, we have verified the applicability of
simple deep learning architectures to the problem
of traffic flow optimization by traffic light timing
control on small and medium-sized traffic
networks. However, for larger-sized networks
further study is needed, probably in the
lines of exploring the results with significantly
larger training times, using the geometric
information of the network and devising data
efficiency improvements.

\clearpage

\bibliographystyle{elsarticle-num-names}
\bibliography{biblio}

\begin{thebibliography}{62}
\bibitem[{Tieleman and Hinton(2012)}]{tieleman2012lecture}
\bibinfo{author}{T.~Tieleman}, \bibinfo{author}{G.~Hinton},
  \bibinfo{title}{Lecture 6.5-rmsprop: Divide the gradient by a running average
  of its recent magnitude}, \bibinfo{journal}{COURSERA: Neural Networks for
  Machine Learning} \bibinfo{volume}{4}~(\bibinfo{number}{2}).
\end{thebibliography}

\end{document}